\documentclass[11pt,a4paper]{article}
\usepackage[hyperref]{ranlp2023}
\usepackage{times}
\usepackage{latexsym}
\usepackage{graphicx}
\usepackage{booktabs}
\usepackage{arydshln}
\usepackage{comment}

\usepackage{microtype}

\aclfinalcopy % Uncomment this line for the final submission
\newcommand{\gr}[1]{}
\newcommand{\jmf}[1]{} %uncomment to remove comments
\newcommand{\nilay}[1]{} %uncomment to remove comments
\newcommand{\cm}[1]{}
\newcommand{\bdk}[1]{}

\title{Does the ``most sinfully decadent cake ever" taste good? Answering Yes/No Questions from Figurative Contexts}

\author{Geetanjali Rakshit and Jeffrey Flanigan \\
Computer Science and Engineering Department\\
UC Santa Cruz \\
\texttt{\{grakshit,jmflanig\}@ucsc.edu} }

\date{}

\begin{document}
\maketitle
\begin{abstract}
Figurative language is commonplace in natural language, and while making communication memorable and creative, can be difficult to understand.
In this work, we investigate the robustness of Question Answering (QA) models on figurative text. Yes/no questions, in particular, are a useful probe of figurative language understanding capabilities of large language models. We propose FigurativeQA, a set of 1000 yes/no questions with figurative and non-figurative contexts, extracted from the domains of restaurant and product reviews. We show that state-of-the-art BERT-based QA models exhibit an average performance drop of up to 15\% points when answering questions from figurative contexts, as compared to non-figurative ones. While models like GPT-3 and ChatGPT are better at handling figurative texts, we show that further performance gains can be achieved by automatically simplifying the figurative contexts into their non-figurative (literal) counterparts. We find that the best overall model is ChatGPT with chain-of-thought prompting to generate non-figurative contexts. Our work provides a promising direction for building more robust QA models with figurative language understanding capabilities.
\end{abstract}

\section{Introduction}
\textit{``Questions are never indiscreet. Answers sometimes are."\begin{flushright} - Oscar Wilde\end{flushright}}

One of the many interesting phenomena occurring in natural language is the presence of figurative language, which, while making communication creative and memorable~\cite{danescu2012you}, may sometimes also prove difficult to understand \cite{zayed2020figure}. This includes (but is not limited to) linguistic constructs such as idioms, similes, metaphors, rhetorical questions, hyperbole, personification, sarcasm, and irony. It may be particularly difficult for non-native speakers to interpret figurative expressions, and phenomena like sarcasm are often missed altogether~\cite{joshi2016cultural}. Given that figurativeness is commonplace in everyday communication~\cite{lakoff2008metaphors}, progress in the field of Natural Language Understanding (NLU) would be incomplete without figurativeness understanding. Consequently, figurative text has been studied in various downstream NLP tasks such as machine translation \cite{dankers2022can}, textual entailment \cite{agerri2008metaphor}, ~\cite{chakrabarty2021figurative},~\cite{liu2022testing} and dialog models \cite{jhamtani2021investigating}, inter-alia. However, to the best of our knowledge, there has not been a systematic study of figurative language understanding capabilities of question answering models.

\begin{figure}[t]
\centering
\begin{tabular}{|l|}
\hline \\
\textit{The cake was described as \textbf{the most sinfully}} \\
\textit{\textbf{decadent ever} .} \\ \\
\textbf{Question}: Did the cake taste good? \\
\textbf{Answer}: Yes \\ \\ \hline
\end{tabular}
\caption{\label{fig-fig-yes-no-question}
To answer the question ``Did the cake taste good?" based on the context, a Question Answering (QA) model needs to be able to correctly infer the meaning of the figurative text ``the most sinfully decadent ever"}
\end{figure}

We focus on yes/no questions for our question answering (QA) task.  Yes/no questions are a good test of figurative language understanding because correctly answering them requires the reader to correctly understand the figurative language.  Extractive QA, on the other hand, is not a good test for figurative language understanding because it does not require actually understanding the figurative language.

For example, if we were to pose the question “How did the cake taste?” from the context “The cake was described as the most sinfully decadent ever.”, an answer such as “sinfully decadent” from an extractive QA model doesn’t really tell us that the model understands the meaning of the figurative text “sinfully decadent”. It simply copies the figurative text and it's up to the reader to infer what the answer means.

However, in order to answer a yes/no question such as “Did the cake taste good?”, a QA model needs to correctly infer that “sinfully decadent” means \textit{rich and delicious}, or in other words, \textit{really good}, and therefore the answer would be \textit{yes}.

%In our setup of figurativeness understanding, we formulate it as a machine comprehension task, where the context is figurative, and answering a question based on the context requires understanding of the figurative expression(s) present in the context. We hypothesize that figurativeness understanding in large-scale pre-trained models can be effectively probed using yes/no questions\jmf{our hypothesis is}\gr{done}. For instance, in \autoref{fig-fig-yes-no-question}, from the context “The cake was described as the most sinfully decadent ever.”, if we were to pose the question “How did the cake taste?”, an answer such as “sinfully decadent” from an extractive Question Answering (QA) model doesn’t really tell us that the model understands the meaning of the figurative text “sinfully decadent”. It simply copies the figurative text and it's up to the reader to infer what the answer means. However, in order to answer a question like “Did the cake taste good?”, a QA model needs to correctly infer that “sinfully decadent” means \textit{rich and delicious}, or in other words, \textit{really good}, and therefore the answer would be \textit{yes}. \bdk{Intro is really good! if removing some italics, I'd remove italics for everything in quotes}

%Figurative language has a limited presence in commonly used Question Answering (QA) datasets such as SQuAD~\cite{rajpurkar2018know} (which is constructed on top of factual text from Wikipedia), but

Despite the lack of attention of figurative language for QA tasks, figurative language is extremely common in some important domains, such as online reviews.  We randomly sampled 100 reviews from the train split of the Yelp Challenge Dataset\footnote{We use the version in Huggingface Datasets (\url{https://huggingface.co/datasets/yelp_review_full}), from the paper~\cite{zhang2015character}}, and observe that at least 60\% of these reviews contain figurative expressions. Users often write strongly-worded reviews, to express highly positive or highly negative opinions about products or services \cite{mohammad2016metaphor}, which tend to contain figurative language.

We show that it can be challenging for existing QA models to draw inferences from figurative text.\jmf{is this still true?} To do this, we present a new dataset, \textbf{\textit{FigurativeQA}}, consisting of 1000 yes/no questions and accompanying figurative and non-figurative contexts constructed from Amazon product reviews \cite{niculae2014brighter} and Yelp restaurant reviews \cite{oraby2017harvesting}.  In~\autoref{fig-figQA-examples}, we show examples from FigurativeQA, in two domains: Amazon product reviews and Yelp restaurant reviews, for both figurative and non-figurative contexts. Each context is accompanied by a question-answer pair, and in the case of figurative contexts, manually constructed and automatically obtained non-figurative versions of the context.

%FigurativeQA contains manually crafted yes/no questions, answering which require understanding of figurative text, when present in the context. We also manually annotate all figurative contexts with their non-figurative (literal) counterparts. We show that for current state-of-the-art models, it is harder to answer questions from figurative contexts. Since manual conversion of figurative to non-figurative text is expensive and time-consuming, we propose prompting GPT-3 \cite{brown2020language} to do this automatically.

\begin{figure*}[hbt!]
\begin{tabular}{llp{0.7\textwidth}}
\hline
\textbf{Split} & \textbf{Source} & \textbf{Example} \\ \hline
Figurative & Amazon & \textbf{Context}: \textit{The album , like almost everything Krush has released , \textbf{slays} .} \\
& & \textbf{Question}: \textit{Is the album good?} \\
& & \textbf{Answer}: \textit{Yes} \\ \hdashline
& & \textbf{Non-fig. version of the context (manual)}: \textit{The album is really good, like most of Krush's work.} \\ 
& & \textbf{Non-fig. version of the context (from GPT-3)}: \textit{The album is really good, like almost everything Krush has released.} \\ \hline
Figurative & Yelp & \textbf{Context}: \textit{Although, the menu items doesnt \cm{do not} \textbf{SCREAM} French cuisine. Most foods looks \cm{look} like you can get at any American place.} \\
& & \textbf{Question}: \textit{Is the menu authentic french?} \\
& & \textbf{Answer}: \textit{No} \\ \hdashline
& & \textbf{Non-fig. context (manual)}: \textit{The menu items aren't typical of French cuisine. Rather, they are common at most American eateries.} \\& & \textbf{Non-fig. context (from GPT-3)}: \textit{Although, the menu items doesn't \cm{do not} look very French. Most foods look like you can get at any American place.} \\ \hline
Non-figurative & Amazon & \textbf{Context}: \textit{Nice ring, but the color is paler than the picture .} \\
& & \textbf{Question}: \textit{Is the ring brightly colored?} \\ 
& & \textbf{Answer}: \textit{No} \\\hline
Non-figurative & Yelp & \textbf{Context}: \textit{the chicken is delicious and so are the ribs} \\
& & \textbf{Question}: \textit{Did the food taste good?} \\ 
& & \textbf{Answer}: \textit{Yes} \\ \hline
\end{tabular}
\caption{\label{fig-figQA-examples}
Examples from the figurative and non-figurative splits of FigurativeQA, from Amazon product reviews and Yelp restaurant reviews. The figurative text fragments within the contexts are shown in bold and italics.\jmf{move to appendix}}
\end{figure*}

We develop a variety of methods for improving QA performance for figurative text.  We prompt powerful LLMs like GPT-3 and ChatGPT to convert figurative contexts to literal as an intermediate step to question answering. We then provide these literal contexts as input to state-of-the-art QA models, resulting in considerable gains in performance. The best performance is achieved by the chain-of-thought prompting method from ChatGPT in a few-shot setting, where the model generates a simplified version of the context and then generates the yes/no answer. We also use these LLMs to generate domain-specific training data to fine-tune models specifically for this task. %We also fine-tune GPT-3 specifically for the task of converting figurative to literal text and find this model to be more effective in getting the right answer in some cases.

%The contributions of this work are the following: \jmf{merge into intro}
%\begin{itemize}
    %\item FigurativeQA, a test set of 1000 yes/no question-answer pairs with figurative and non-figurative contexts. For the 200 figurative contexts, we also provide manually created literal contexts for comparison.
    %\item We show that it is harder to answer questions from figurative contexts for models trained on QA data with non-figurative contexts, and that manually changing the figurative context to a meaning-preserving non-figurative version improves performance.
    %\item We propose a method to use GPT-3 to automatically produce non-figurative contexts from figurative ones, and demonstrate that it improves our QA system on the FigurativeQA dataset.
%\end{itemize}

The outline of the paper is as follows: after reviewing related work (\S\ref{sec:related_work}), we introduce our new QA dataset for figurative language, FigurativeQA, in (\S\ref{sec:dataset}). We report baseline QA performance on FigurativeQA and introduce a method for simplifying figurative language to non-figurative by prompting GPT-3 and ChatGPT, which we use to improve our baseline QA models (\S\ref{sec:expts_part1}, \ref{sec:expts_part2}, \ref{sec:expts_part3}). We report our experiments with chain-of-thought prompting in \S\ref{sec:expts_part4}. We prompt ChatGPT to generate in-domain training data for figurative question answering (\S\ref{sec:expts_part5}). We finally conclude in (\S\ref{sec:conclusion}). The FigurativeQA dataset can be accessed at \url{https://github.com/geetanjali-rakshit/figurativeQA}.

\section{Related Work}
\label{sec:related_work}
Figurative language has been a difficult problem for many natural language processing (NLP) applications. A number of computational approaches have been proposed to study their occurrence in text \cite{veale2016metaphor,qadir2016automatically,kordoni2018beyond,mao2018word, zhang2017asking, troiano2018computational}, including generation of figurative language \cite{chakrabarty2020generating,zhou2021solving}. 
%The challenge of handling figurative language has been studied in downstream tasks such as machine translation \cite{mao2018word,dankers2022can}, recognizing textual entailment \cite{chakrabarty2021figurative}, sentiment analysis \cite{qadir2015learning}, among others. ~\citet{chakrabarty2021figurative} investigate the robustness of state-of-the-art entailment models on figurative examples on test sets for similes, metaphors, and irony. \citet{chakrabarty2022s} test figurative language understanding in pre-trained language models by evaluating continuation of text in narratives, while~\citet{liu2022testing} investigate non-literal reasoning capabilities of language models on a Winograd-style non-literal language understanding task.

The idea of converting metaphors to their literal counterparts has been previously explored for machine translation by \citet{mao2018word}, where metaphors in English text are first identified and then converted to a literal version by using word embeddings and WordNet, before doing machine translation into Chinese. In dialog systems, a similar approach was employed by \citet{jhamtani2021investigating}, where idioms and metaphors in utterances are converted to literal versions using a dictionary lookup-based method. Our work is closest to \citet{jhamtani2021investigating}, except that we explore the robustness of QA systems in a machine comprehension setup, instead of dialog models, to figurative language, which, to the best of our knowledge, is a first. Our automatic approach to creating rephrased non-figurative versions of figurative text is done using pre-trained language models, rather than rule-based  methods which have been shown to be error-prone~\cite{jhamtani2021investigating}. In a concurrent work, \citet{chakrabarty2022flute} have also done prompting on GPT-3 to create their figurative NLI dataset, FLUTE, as well as obtain an explanation of the NLI labels in this dataset. 

To our knowledge, there are no QA datasets specifically designed for figurative language understanding, but some existing QA datasets do contain figurative language. The FriendsQA dataset~\cite{yang2019friendsqa} is a dialog-based QA dataset constructed from dialogs from the TV series Friends. While it does contain metaphors and sarcasm, the focus of the dataset is not figurative language, and it is not ideal for testing figurative language understanding as it is unclear how much of the dataset is figurative. The dialog nature of the dataset further contributes to making it challenging and complicates studying the effect of figurativeness. Another dataset that requires figurative language understanding is the RiddleSense dataset~\cite{lin2021riddlesense}, which comprises of riddles, but unlike ours, it's modeled as an open-domain QA task rather than a machine comprehension task. \citet{parde2018automatically} show that questions about novel metaphors from literature are judged to be deeper than non-metaphorical or non-conventional metaphors by humans, but their focus is on generating deep questions rather than testing the robustness of QA models. \citet{dankincan} construct yes/no questions using templates to detect the presence of metaphors in a few-shot setting. 

\section{FigurativeQA Dataset}
\label{sec:dataset}
\jmf{shorten/move non-critical parts to appendix}
The contexts in FigurativeQA comes from two sources: Amazon product reviews \cite{niculae2014brighter}, and Yelp restaurant reviews \cite{oraby2017harvesting}. We extract both figurative and non-figurative contexts from each source. We manually construct yes/no questions and answers on top of these contexts.~\autoref{fig-figQA-examples} shows examples from FigurativeQA. The data statistics from each source (Amazon and Yelp) and each split (figurative and non-figurative) are summarized in ~\autoref{tab-stats-yes-no-questions}. 

\begin{table}[htbp!]
\centering
\begin{tabular}{c|cc|cc}
\hline
 & \multicolumn{2}{c}{\textbf{Amazon}} & \multicolumn{2}{|c}{\textbf{Yelp}} \\ 
 & \textbf{Fig.} & \textbf{Non-fig.} & \textbf{Fig.} & \textbf{Non-fig.}\\ \hline
\textbf{Yes} & 77 & 76 & 174 & 175 \\
\textbf{No}  & 73 & 74 & 176 & 175 \\ \hline
\textbf{Total} & 150 & 150 & 350 & 350 \\ \hline
\end{tabular}
\caption{\label{tab-stats-yes-no-questions}
Distribution of yes/no questions from Amazon product reviews and Yelp restaurant reviews for figurative and non-figurative contexts
}
\end{table}

%\begin{table}[hbt!]
%\centering
%\begin{tabular}{lllll}
%\hline
%& \textbf{avg.} & \textbf{category} & \textbf{Yes} & \textbf{No}\\ 
% & \textbf{context} & &  & \\
% & \textbf{length} & &  & \\ \hline
%\textbf{fig.} & 9 & \textbf{Amazon} & 52 & 48 \\
%& 16 & \textbf{Yelp} & 50 & 50 \\ \hline
%\textbf{non-fig.} & 10 & \textbf{Amazon} & 50 & 50 %\\
%& 14 & \textbf{Yelp} & 49 & 51 \\ \hline
%\end{tabular}
%\caption{\label{tab-stats-yes-no-questions}
%Number of yes-no questions from Amazon product reviews and Yelp restaurant reviews for figurative and non-figurative contexts, and the average length of context (number of words)
%}
%\end{table}

\begin{figure*}[hbtp!]
\begin{tabular}{lp{0.65\textwidth}l}
\hline
\textbf{Split} & \textbf{Context} & \textbf{fig. construct} \\ \hline
\textbf{Amazon} & \textit{The books are \textbf{like potato chips} - you \textbf{can't eat just one} .}  & \textit{simile, idiom} \\ 
& \textit{So when my laptop battery puffed up \textbf{like a balloon} , I dreaded paying the cost of replacement .}  & \textit{simile, hyperbole} \\ 
& \textit{Really , this novel feels more \textbf{like a footnote} to the series whereas The Gunslinger was a novel that \textbf{stood extremely well on its own} .}  & \textit{simile, idiom} \\
& \textit{These horrible recordings \textbf{contain treasure more precious than gold}.}  & \textit{simile, sarcasm} \\ \hline
\textbf{Yelp} & \textit{i had the chicken fajitas , which came with a giant flour tortilla that was \textbf{as hot as hades} .} & \textit{simile, hyperbole}\\ 
% & \textit{the chicken was super dry and overcooked lol i had to try to soak it in as much dressing as i can \textbf{fish from the bottom} of the bowl} . & idiom\\ \hline
& \textit{the cheese was scarce as was the meat , and the taste was nothing to \textbf{write home about} .} & \textit{idiom}\\ 
& \textit{i ate as much as i could because truly , underneath the \textbf{salt mine} on my plate , was some damn fine corned beef hash !} & \textit{metaphor, hyperbole}\\ \hline 
\end{tabular}
\caption{\label{fig-fig-constructs-examples}
Examples of figurative constructs observed in the Amazon and Yelp datasets. The figurative text fragments within the contexts are shown in bold and italics. In case of multiple labels occurring in the same context, the first bold fragment corresponds to the first label, and so on. In some cases, the same text fragment may have multiple labels (as in row 2)\jmf{move to appendix}}
\end{figure*}

\begin{table}[hbt!]
\centering
\begin{tabular}{lcc}
\hline
\textbf{Figurative Construct} & \textbf{Amazon} & \textbf{Yelp}\\ \hline 
\textbf{Simile} & 91 & 70 \\ 
\textbf{Metaphor} & 20 & 35 \\ 
\textbf{Hyperbole} & 18 & 44 \\
\textbf{Idiom} & 15 & 2 \\
\textbf{Sarcasm} & 2 & 20 \\\hline
\end{tabular}
\caption{\label{tab-fig-constructs-distribution}
Distribution of occurrences of various kinds of figurative constructs in a random sample of 100 contexts from  Amazon and Yelp each. It is common for a context to contain multiple figurative expressions, so these do not add up to 100\% (refer to~\autoref{fig-fig-constructs-examples} for examples).
}
\end{table}

%Additionally, we manually construct a non-figurative version of the figurative contexts, to be able to determine how much figurative language affects the model performance (\S\ref{sec:experiments}).

For the Amazon part of FigurativeQA, we use \citet{niculae2014brighter}'s dataset of figurative comparisons. %extracted using comparator patterns (such as \textit{"like"}, \textit{"as"}, or \textit{"than"}) from Amazon product reviews. %This dataset comes with 3 sets of figurativeness scores (on a scale of 1 to 4) done on Amazon Mechanical Turk (with scores of 1–2 binned as literal and 3–4 as figurative). 
Of the 1260 comparisons in this dataset, we extract instances where all 3 annotators are in agreement about figurativeness (i.e., average figurativeness score of greater than 3). We then randomly pick 150 examples to form the set of figurative contexts. From the examples with a low average figurativess score, we select 150 examples to form the set of non-figurative contexts.

For the Yelp part of the dataset, the contexts are sourced from \cite{oraby2017harvesting}'s NLG dataset for the restaurant domain. %These come with sentiment annotation (1-2 rating for negative, 3 for neutral and 4-5 for positive), but there is no figurativeness information. 
Since highly positive or highly negative reviews are more likely to contain figurative language, we extract these first, and then, similar to \cite{niculae2014brighter}, use comparator expressions to get a set of reviews likely to be rich in figurative content. We then manually examine these reviews to annotate 350 examples of figurative contexts and non-figurative contexts, each. 

The figurative contexts from FigurativeQA tend to contain more \textit{similes}, since comparator patterns (\textit{``like"}, \textit{``as"}, or \textit{``than"}) were used to extract the text. However, we observe that many of these examples also contain other kinds of figurative constructs such as metaphor, idiom, hyperbole, sarcasm, etc.~\autoref{tab-fig-constructs-distribution} shows the number of occurrences of various kinds of figurative constructs that we observe in a random set of 100 figurative contexts, each from Amazon and Yelp in FigurativeQA.~\cite{oraby2017harvesting} note that one of the most prominent characteristics of restaurant reviews in the Yelp corpus is the prevalence of hyperbole, which we also observe in this sample. A context may contain multiple figurative elements, coming from different text fragments within the context. Also, in some cases, the same text fragment may denote multiple kinds of figurative constructs. In ~\autoref{fig-fig-constructs-examples}, we show some examples of various kinds of figurative constructs occurring in FigurativeQA. 

For each context in FigurativeQA, we construct a yes/no question. For the figurative contexts, we make sure to pose a question such that answering it would require an understanding of the figurative text present in the context. For the non-figurative contexts, we construct questions similar to the ones for the figurative contexts. Additionally, for the figurative contexts extracted from Amazon and Yelp, we manually create non-figurative counterparts that preserve the meaning and overall content. 

\subsection{Inter-annotator Agreement}
\jmf{move to appendix}
Annotations for all the data in FigurativeQA (figurativeness scores for the examples from Yelp, construction of question-answer pairs, manual conversion of figurative contexts to non-figurative) were done by an in-house-trained graduate-student annotator.
To assess the quality of figurative and non-figurative contexts for the Yelp contexts, we perform a second round of annotations with another trained annotator on a random sample of 50 contexts. This resulted in an inter-annotator agreement of 0.72 on figurativeness, calculated by Cohen's $\kappa$.

Similarly, to assess the overall quality of FigurativeQA, we randomly sample 50 figurative contexts for double annotation, which gives an additional set of annotations for the answers to the questions. The inter-annotator agreement on the answers was found to be 0.96, calculated by Cohen's $\kappa$. To validate the effectiveness of the questions for figurativeness comprehension, we also asked the annotators to indicate if answering the question required them to understand figurative text fragments present in the context. In the random sample of 50, in 49 cases the annotators were in agreement that this was indeed the case. 

\begin{comment}
\section{Experiments}
\label{sec:experiments}
\gr{first draft}
%\jmf{expand upon this (method and discuss the results)}
We perform experiments on FigurativeQA to answer the following questions:

\begin{itemize}
    \item Do QA models find answering questions from figurative contexts harder? (\S\ref{sec:expts_part1})
    \item Can prompting GPT-3 help simplify figurative contexts? (\S\ref{sec:expts_part2}) 
    \item Can GPT-3 generated simplified text help QA performance? (\S\ref{sec:expts_part3})
    \item How much does the prompting method help with handling figurativeness? (\S\ref{sec:expts_part4})
\end{itemize}
We describe these experiments and our conclusions in the following subsections. 
\end{comment}

\section{Do QA models find answering questions from figurative contexts harder?} \label{sec:expts_part1}
%% add topic sentences in each paragraph
%% start with high-level description of the paragraph content
Using FigurativeQA as a test set, we show that current models struggle to do well on figurative text compared to literal ones\jmf{update to something ``BERT-based fine-tuned models''}. We use a RoBERTa-based \cite{liu2019roberta} QA model fine-tuned on BoolQ to show this. The BoolQ dataset~\cite{clark2019boolq} consists of yes/no questions from the Natural Questions dataset. We use the training split of BoolQ containing 9,427 examples to fine-tune RoBERTa-base and report its performance on FigurativeQA in~\autoref{tab-baseline}. We find that the RoBERTa QA model performs poorly on the figurative contexts compared to the non-figurative contexts, with a drop in performance of $\sim$8.5\% points for Amazon, and $\sim$23\% points for Yelp. We observe that switching the figurative contexts for their manually created non-figurative counterparts shoots these numbers up in both cases, by $\sim$10\% points and $\sim$23\% points, for Amazon and Yelp, respectively. 
More powerful models like ChatGPT (in a few-shot setting) perform significantly better on figurative contexts, but still don't match the results on non-figurative versions of the contexts. This indicates that the conversion of figurative language to non-figurative language may help improve QA performance.

\cm{The table is out of the boundary, you can add resize box around the table(I added it for you) }
\begin{table}[hbt!]
\centering\resizebox{\columnwidth}{!}{
\begin{tabular}{lll}
\hline
& \textbf{Amazon} & \textbf{Yelp} \\ \hline
\textbf{RoBERTa-BoolQ} & & \\
Fig (Original) & 83.4 $\pm$ 0.7 & 66.8 $\pm$ 1.4 \\
Fig (manual non-fig) & \textbf{93.5 $\pm$ 1.1$^*$} & \textbf{90.0 $\pm$ 1.4$^*$} \\
Non-fig (Original) & 92.0 $\pm$ 1.4 & 89.8 $\pm$ 1.7 \\
\hline 
\textbf{ChatGPT(few-shot)} & & \\
Fig (Original) & 92.6$\pm$1.1  & 80.6$\pm$0.7\\
Fig (manual non-fig) & \textbf{93.8 $\pm$0.3$^*$} & \textbf{83.3$\pm$1.6$^*$}  \\
Non-fig (Original) & $93.5\pm0.3^*$ & $88.7\pm1.8^*$ \\
\hline 
\end{tabular}}
\caption{\label{tab-baseline}
Accuracy of RoBERTa-base fine-tuned on BoolQ, and ChatGPT (few-shot), on the figurative split, manually created non-figurative version of the figurative split, and non-figurative split of FigurativeQA. (We reran experiments 1000 times with bootstrap resampling. The numbers reported are the mean and std-dev. $^*$ denotes statistically significant results, with $p < 0.05$ calculated using the Wilcoxon signed-rank test. The numbers in \textbf{bold} are the best results.)}
\end{table}

\section{Can prompting or finetuning LLMs help simplify figurative contexts?} \label{sec:expts_part2}
We posit that answering questions from figurative contexts is harder, and that simplifying the figurative context into its literal/non-figurative version improves QA performance. However, since the task of manually converting figurative text to non-figurative is expensive and time-consuming, we propose to do this automatically by prompting GPT-3 \cite{brown2020language} in two ways. First, we use GPT-3 (da-vinci-003) and ChatGPT in a few-shot setting to generate non-figurative/literal versions of all the figurative contexts in FigurativeQA.\footnote{The experiments for this method to convert figurative text to non-figurative were performed by running API calls to the OpenAI da-vinci model. For each context, this took less than 1 second, for a total of less than 18 min and cost less than 8 USD for the entire dataset.} We also used a similar approach to prompt ChatGPT. Please refer to Appendix A for model details and the prompts used. Second, we use a trained version of GPT-3 (da-vinci-002) fine-tuned specifically for the task of converting figurative to literal text. 

As an intrinsic evaluation of the effectiveness of our prompting method, we manually evaluate the correctness of the non-figurative/literal contexts generated by prompting GPT-3 on a random sample of 100 instances each, from Amazon and Yelp in FigurativeQA. We label each generated literal version as either \textbf{``correct"}, where none of the figurative expressions are present but the meaning is preserved, or \textbf{``incorrect"} where the generated output is the same/similar to the original context or the meaning has changed. Please note that this is a rather strict evaluation of correctness, as in some cases, some of the figurative text fragments present in the context is converted to literal, while the context may still be left with some amount of figurativeness (possibly arising from multiple figurative text fragments present in the context).~\autoref{tab-results-gpt3-outputs} shows the results from the manual evaluation of the GPT-3 and ChatGPT outputs. We observe that these models are pretty good at converting figurative language in FigurativeQA to literal, with nearly 89\% and 81\% of the outputs from GPT-3 judged to be correct in Amazon and Yelp, respectively, and 92\% and 88\% for ChatGPT. In~\autoref{fig-gpt3-examples}, we show examples of non-figurative text generated from GPT-3 and ChatGPT. 

\begin{table}[hbt!]
\centering
\begin{tabular}{lcc}
\hline
 & \textbf{Amazon} & \textbf{Yelp}\\ \hline 
GPT-3 & 89\% & 81\% \\ \hline
ChatGPT & \textbf{92\%} & \textbf{88\%}  \\ \hline
Finetuned GPT-3 & 80\% & 77\% \\ \hline
\end{tabular}
\caption{\label{tab-results-gpt3-outputs}
Evaluation of non-figurative outputs from GPT-3 and ChatGPT, showing the percentage of generated outputs that do not contain figurative expressions, but preserve the original meaning of the figurative context.\jmf{add standard deviation}\gr{skipped}
}
\end{table}

%We observe that for the figurative contexts from Yelp, GPT-3 struggles to generate  a correct non-figurative version and often outputs the same text as the input\jmf{perhaps speculate on why this is the case (why Yelp is harder for GPT-3)}.  

\begin{figure*}[hbt!]
\begin{tabular}{lp{0.9\textwidth}}
\hline
\textbf{Amazon} & \textbf{Figurative Context}: \textit{However , the obvious problem with Eragon hits \textbf{like a brick wall} .} \\
& \textbf{[CORRECT] Non-fig. version from GPT-3}: However, the obvious problem with Eragon is glaringly obvious. \\ 
& \textbf{[CORRECT] Non-fig. version from ChatGPT}: However, the obvious problem with Eragon is very clear. \\ 
\hline
& \textbf{Figurative Context}: \textit{Not a storybook , by any means , this one is more \textbf{like a visit to the zoo} .} \\
& \textbf{[INCORRECT] Non-fig. version from GPT-3}: \textit{Not a fairytale, by any means, this one is more like a visit to the zoo.} \\
& \textbf{[INCORRECT] Non-fig. version from ChatGPT}: \textit{Not a fairytale, by any means, this one is more like a visit to the zoo.}  \\ \hline \hline
\textbf{Yelp} & \textbf{Figurative Context}: \textit{this is as authentic thai \textbf{as much as imitation crab is authentic crab} .} \\
& \textbf{[INCORRECT] Non-fig. version (from GPT-3)}: \textit{this is as authentic thai as much as imitation crab is genuine crab.} \\
& \textbf{[CORRECT] Non-fig. version from ChatGPT}: \textit{This is not authentic Thai, just as imitation crab is not authentic crab.} \\ \hline 
& \textbf{Figurative Context}: \textit{the same thing with the steak and potatoes , it was almost as if they tried to \textbf{decorate the plate with salt} .} \\
& \textbf{[CORRECT] Non-fig. version from GPT-3}: \textit{The steak and potatoes were heavily salted, as if they were trying to make the plate look more appealing.} \\ 
& \textbf{[CORRECT] Non-fig. version from ChatGPT}: \textit{The steak and potatoes were oversalted and appeared to be more about presentation than taste.} \\ \hline

\end{tabular}
\caption{\label{fig-gpt3-examples}
Examples of non-figurative contexts generated from GPT-3, for Amazon and Yelp. The figurative text fragments within the contexts are shown in \textbf{bold} and \textit{italics}.}
\end{figure*}

We next explore using a fine-tuned version of GPT-3 to generate literal versions of figurative texts.~\citet{chakrabarty2022flute} propose the FLUTE dataset for Natural Language Inference (NLI), which has 9,000 figurative NLI instances, and explanations for the NLI labels.
We extract the premise-hypothesis pairs with the label \textit{``entailment"} from the training split of FLUTE to fine-tune GPT-3 (3,182 examples in total). We used the \textit{davinci} model from OpenAI as the base model and fine-tuned for 4 epochs, with all default settings. We didn't perform any hyper-parameter tuning.\footnote{To fine-tune GPT-3 on the FLUTE dataset, it cost about 15 USD and took 62 minutes.}~\autoref{tab-results-gpt3-outputs} (row 3) shows the results from manual evaluation of the fine-tuned GPT-3 outputs. \jmf{rewrite to be more clear: we fine-tune a model on the FLUTE dataset by...}

\section{Can automatically generated non-figurative text improve QA performance?} \label{sec:expts_part3}

We observed that ChatGPT has a much stronger performance on FigurativeQA than the baseline model of RoBERTa finetuned on BoolQ (\autoref{sec:expts_part1}), and both of these models do better on non-figurative texts. We showed that both GPT-3 and ChatGPT can be effectively used to simplify figurative text into their non-figurative counterparts (\autoref{sec:expts_part2}). We next experiment with simplifying contexts to boost QA performance. As competitive baselines, we also report zero-shot and few-shot QA performance\footnote{Please refer to Appendix B for details about prompting GPT-3 and ChatGPT as a QA system.} of GPT-3 and ChatGPT in~\autoref{tab-overall-results}.
Besides the RoBERTa-finetuned-on-BoolQ baseline (previously described in~\autoref{sec:expts_part1}, we also fine-tune GPT-3 on the training split of BoolQ. For fine-tuning GPT-3, we used the \textit{davinci} model from OpenAI as the base model and fine-tuned for 4 epochs, with all default settings. We didn't perform any additional hyper-parameter tuning.

\begin{table*}[t!]
\centering
\begin{tabular}{|l|cc|cc|cc|}
\hline
 & \multicolumn{2}{c}{\textbf{Fig.}} & \multicolumn{2}{|c}{\textbf{Non-fig.}} & \multicolumn{2}{|c|}{\textbf{Overall}} \\ 
 & \textbf{Amazon} & \textbf{Yelp} & \textbf{Amazon} & \textbf{Yelp} & \textbf{Amazon} & \textbf{Yelp}\\
\hline
\hline
\multicolumn{7}{|l|}{\textbf{Zero-Shot}}\\
\hline
GPT-3 (zero) & 71.9$\pm$1.2 & 60.2$\pm$3.2 & 88.7$\pm$0.9 & 86.0$\pm$2.2 & 80.3$\pm$1.1 & 73.1$\pm$2.1\\
\hline
ChatGPT (zero) & 91.0$\pm$0.7 & 87.4$\pm$2.6 & 93.0$\pm$0.3 & 88.6$\pm$2.4 & 92.0$\pm$0.5 & 88.0$\pm$2.3\\
\hline
\hline
\multicolumn{7}{|l|}{\textbf{Few-Shot}}\\
\hline
GPT-3 (few) & 85.7$\pm$1.8 & 64.1$\pm$3.7 & 90.2$\pm$0.8 & 88.3$\pm$1.9 & 88.0$\pm$1.1 & 76.2$\pm$2.7\\
\hline
ChatGPT (few) & 92.6$\pm$1.1 & 80.6$\pm$0.7 & 93.5$\pm$0.3 & $88.7\pm1.8$& 93.0$\pm$0.7 & 84.7$\pm$1.1 \\ \hline
\hline
\multicolumn{7}{|l|}{\textbf{Supervised}}\\
\hline
RoBERTa & 83.2$\pm$1.1 & 66.8$\pm$2.6 & 92.2$\pm$1.4 & 89.6$\pm$1.7 & 87.7$\pm$0.9 & 78.2$\pm$1.6\\
\hline
GPT-3-BoolQ & 86.3$\pm$2.1 & 69.2$\pm$3.8 & 88.7$\pm$0.9 & 86.5$\pm$1.2 & 87.5$\pm$1.4 & 77.9$\pm$2.2\\ 
\hline
RoBERTa & \textbf{95.3$\pm$0.5} & \textbf{92.3$\pm$0.7} & 95.8$\pm$1.2 & 90.8$\pm$1.6 & 95.5$\pm$0.7 & \textbf{91.5$\pm$0.9}\\
+synthetic &  &  &  &  &  &  \\
\hline
\hline
\multicolumn{7}{|l|}{\textbf{Simplified Contexts}}\\
\hline
GPT-3+ & $86.5\pm1.1$ & $73.4\pm1.7$ & $92.4\pm1.1$ & $89.4\pm1.7$ & $89.5\pm3.2$ & $81.5\pm1.2$\\
RoBERTa &  &  &  &  &  &  \\
\hline
GPT-3-FLUTE & 88$\pm$0.7 & 69.4$\pm$2.1 & 92.0$\pm$0.4 & 89.5$\pm$1.2 & $90.0\pm1.4^*$ & $79.4\pm2.3^*$\\
+RoBERTa &  &  &  &  &  &  \\
\hline
ChatGPT+ & 88.7$\pm$1.6 & 75.3$\pm$3.5 & 92.2$\pm$1.1 & 89.5$\pm$2.1 & 90.5$\pm$1.2 & 82.4$\pm$3.2\\
RoBERTa &  &  &  &  &  &  \\
\hline
ChatGPT+ & 89.3$\pm$0.8 & 91.0$\pm$0.3 & 95.7$\pm$0.7 & 91.2$\pm$0.2 & 92.5$\pm$0.4 & 91.1$\pm$0.3\\
ChatGPT (few) &  &  &  &  &  &  \\
\hline
ChatGPT+CoT & 94.7$\pm$0.3 & 91.6$\pm$1.2 & \textbf{96.4$\pm$1.1} & \textbf{91.4$\pm$0.7} & \textbf{95.6$\pm$0.9} & \textbf{91.5$\pm$1.1}  \\ \hline
\end{tabular}
\caption{\label{tab-overall-results}
QA accuracy on FigurativeQA. (We reran experiments 1000 times with bootstrap resampling. The numbers reported are the mean and std-dev. $^*$ denotes results that are not statistically significant compared to the best results, with $p < 0.05$ calculated using the Wilcoxon signed-rank test. The numbers in \textbf{bold} are the best results.) GPT-3 finetuned models use da-vinci-002 as the base model.
}
\end{table*}

%The performance on FigQA is summarized in Tables~\ref{tab-baseline} and~\ref{tab-results}. We find that the RoBERTa QA model performs poorly on the figurative contexts compared to the non-figurative contexts, and that manually changing the figurative language to non-figurative language improves performance. This indicates that automatic conversion of figurative language to non-figurative language may improve performance. 

%To improve upon the baseline model, we pass the automatic non-figurative contexts from GPT-3 (\S\ref{sec:method}) to our RoBERTa-base model. We find that this procedure improves the performance on figurative language split, and has no effect on the non-figurative language split, and improves overall performance on FigQA.
%As an additional comparison, we also prompt GPT-3 as a QA model and report its performance on FigQA.

%We categorize these outputs into the following:
%\begin{itemize} 
%\item \textit{correct}, literal versions of the figurative text, such that %they simplify the text but preserve the meaning
%\item \textit{incorrect}, such that GPT-3 rephrases the figurative text to some extent but the meaning is no longer preserved. 
%\item \textit{same}, such that GPT-3 produces the exact same output as the input.
%\end{itemize}

In our experiments, we do not require knowing which contexts are figurative and which are non-figurative.  We simply input both figurative and non-figurative contexts to the LLM to simplify any figurative language that is present, regardless if the context actually contains figurative language.
%In our approach of simplifying contexts provided to the QA models, we use the GPT-3 and ChatGPT generated literal versions of all the contexts (figurative and non-figurative) in FigurativeQA as the input contexts to the baseline RoBERTa-finetuned-on-BoolQ model. 
In ~\autoref{tab-overall-results}, we show that this method exhibits significant gains over the baseline RoBERTa model. We also report the performance of using GPT-3-finetuned-FLUTE as input to the RoBERTa baseline.

%In our method "ours (fine-tuned)" where we use the GPT-3 generated literal versions from GPT-3 fine-tuned on the FLUTE dataset, as input context to the baseline RoBERTa model. Our method is observed to have the best results, compared to the baseline models (\autoref{tab-overall-results}).\jmf{make it clear that we don't tell the model if the text is figurative or not when simplifying it}

%In Figures~\ref{fig-gpt3-corrected-examples} and \ref{fig-gpt3-finetuned-corrected-examples}, we show examples where our approach improves upon the baseline model with the help of literal contexts generated from zero-shot GPT-3 and GPT-3 fine-tuned on the FLUTE, respectively.

\section{Can we use chain-of-thought prompting for improving QA performance on FigurativeQA?} \label{sec:expts_part4}

~\citet{wei2022chain} have shown chain-of-thought prompting in Large Language Models (LLMs) to be effective for solving tasks requiring complex reasoning. Since understanding figurative language often requires implicit reasoning, we investigate the effect of applying chain-of-thought prompting for FigurativeQA using ChatGPT. (Our few-shot prompt for the chain-of-thought method is described in Appendix C.) This approach gives us the highest overall accuracy on FigurativeQA (\autoref{tab-overall-results}).

\section{Can we prompt LLMs to generate training data for FigurativeQA?} \label{sec:expts_part5}
Due to the lack of training data for question answering with figurative contexts, our supervised models are all finetuned on BoolQ.  We hypothesize that adding synthetically generated QA pairs for this task will improve performance of the fine-tuned models.
We prompt ChatGPT to generate synthetic training data (we tried a variety of prompts -- refer to Appendix D for the prompt used).
We use contexts from both Amazon and Yelp domains to generate question answer pairs from ChatGPT. For the Amazon contexts, we randomly sample reviews from 4 categories (Books, Electronics, Jewelry and Digital Music) from Amazon Product reviews from~\cite{mcauley2013hidden}. From these reviews, we extract sentences containing comparator patterns (``like'', ``as'', ``than'') and use them as contexts, as they are more likely to contain figurative expressions. For the Yelp contexts, we extract sentences from~\cite{oraby2017harvesting}'s NLG dataset also containing the same comparator patterns, but not already included in FigurativeQA. (Refer to Appendix E for statistics of the data generated for training.)

We find that further finetuning RoBERTa-finetuned-on-BoolQ on synthetic QA data generated from ChatGPT yields the best performance on the figurative split of both Amazon and Yelp (\autoref{tab-overall-results}).

\section{How much does the prompting method help with handling figurativeness?} \label{sec:expts_part6}

\begin{figure}[hbt!]
\includegraphics[width=0.5\textwidth]{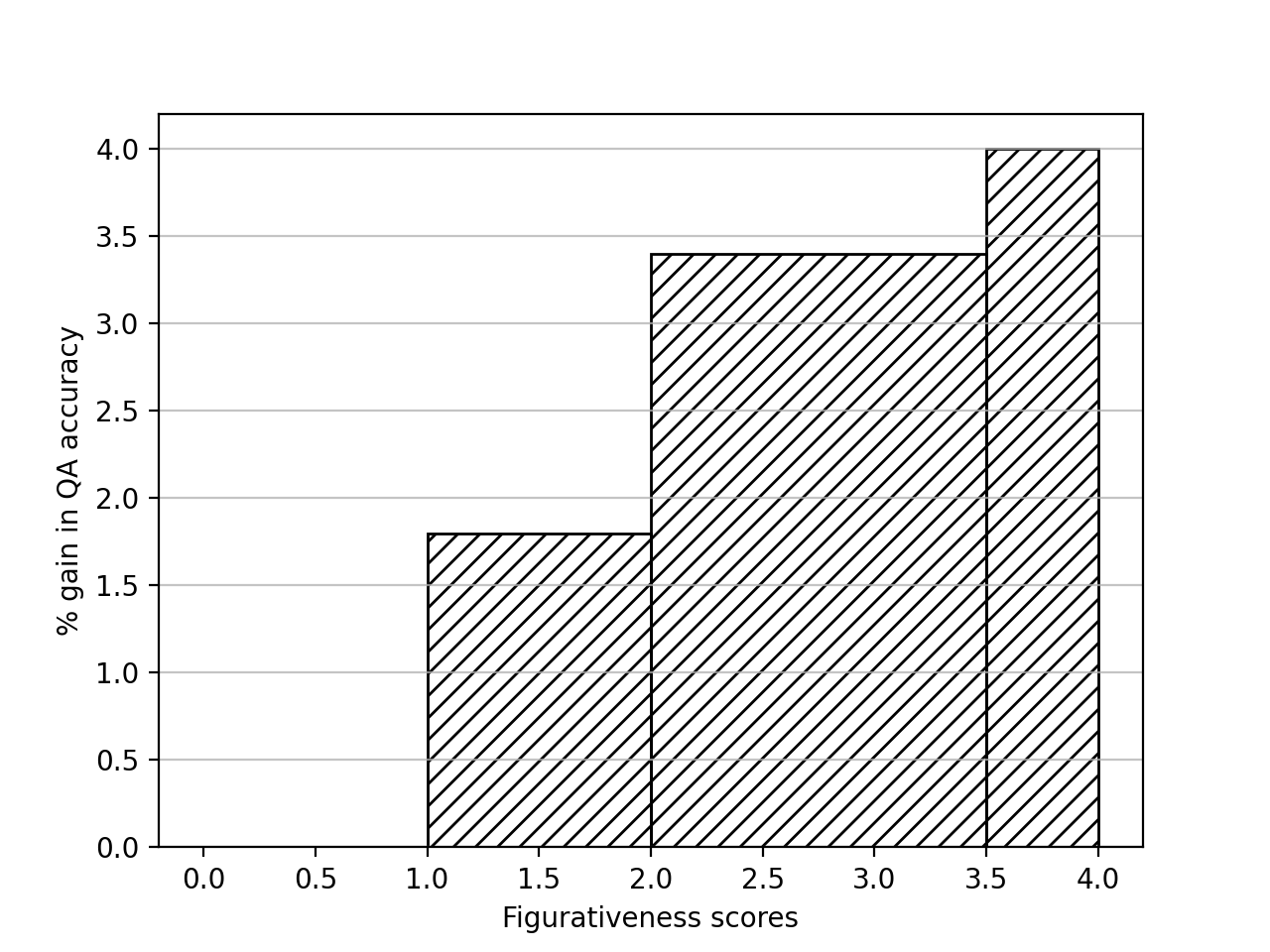}
\caption{Figurativenss Vs Accuracy for the instances from Amazon reviews}
\label{fig-fig_vs_acc}
\end{figure}

%The conversion of figurative text to non-figurative may not always be a straightforward process, but it has been shown to lead to performance gains. 
Our experiments show that the process of converting figurative text into literal by prompting GPT-3 may effectively be used for improving question answering performance. We also study the effect of our method on the degree of figurativeness present in the text. The Amazon reviews data from ~\cite{niculae2014brighter} comes labeled with figurativeness scores of 1-4, with 3 sets of annotations. Using the average figurativeness scores, we bin the Amazon reviews examples in FigurativeQA into 4 splits, and compute the improvement in QA performance when using our method over the baseline. As evident from~\autoref{fig-fig_vs_acc}, the more figurative examples show a higher gain in QA performance.

%\section{QA with Generated Non-Figurative Language}

%\jmf{todo:describe our method that improves upon the baseline (prompting gpt3 to get non-figurative versions)}

%\subsection{Experiments}

%\section{Discussion}
%\jmf{integrate into rest of paper}

\section{Conclusion and Future Work}
\label{sec:conclusion}

We demonstrate that current QA models have reduced accuracy when answering questions from figurative contexts compared to literal ones. This indicates the need for QA models that are robust to figurative language. By manually creating non-figurative versions of these contexts, we observe a significant improvement in performance. To automate this approach, we propose a method of prompting GPT-3 to produce simplified, non-figurative contexts, which yields significant performance gains over the baseline. Chain-of-thought prompting using ChatGPT has the best overall performance on FigurativeQA.  
%In the future, we would like to do a fine-grained analysis of QA performance on different kinds of figurative constructs, including similes, metaphors, irony, idioms, rhetorical questions, hyperbole, personification, etc.
We hope that our method and dataset will spur more research into question answering with figurative language.

\section{Acknowledgments}
This research was supported in part by a research gift from eBay Inc.

\section*{Limitations}
Our dataset contains the specific domains of Amazon and Yelp reviews, which is English-only, and results and conclusions may not generalize to other domains or languages. The text generated by prompting GPT-3 may sometimes produce text that is not faithful to the original figurative text. 

% Entries for the entire Anthology, followed by custom entries
\bibliography{anthology,ranlp2023}
\bibliographystyle{acl_natbib}

\appendix

\section{Appendix A: Prompts for GPT-3 and ChatGPT for simplifying figurative text}
\label{sec:appendixA}
For GPT-3, we use the da-vinci-003 model with temperature set to 0 and max-length set to 100. For ChatGPT, we use gpt-3.5-turbo. In each case, we use a prompt with 5 examples, as shown in~\autoref{fig-gpt3-prompt}.

\begin{figure}[htbp!]
\begin{tabular}{p{0.45\textwidth}}
\hline
For the following inputs, if the text contains figurative language, convert it to a literal version. Otherwise, output the same text as the input. \\ \\

Input: It’s inevitable. Their love was built on sand and this is why their marriage has landed on the rocks. \\
Output: It’s inevitable. Their love was unstable and this is why their marriage has failed. \\ \\

Input: The weather forecast predicted a heatwave this week across most of the country. \\
Output: The weather forecast predicted a heatwave this week across most of the country. \\ \\

Input: During the heatwave, the entire house was like a furnace. \\
Output: During the heatwave, the entire house was uncomfortably hot. \\ \\

Input: The brisket is nothing to write home about. \\
Output: There is nothing particularly remarkable about the brisket. \\ \\

Input: The fries were served cold. \\
Output: The fries were served cold. \\ \\

Input: The lamb had a melt in the mouth texture. \\
Output: The lamb was soft and well-cooked. \\ \\

Input: The adapter worked like a charm. \\
Output: The adapter worked perfectly. \\
\hline
\end{tabular}
\caption{\label{fig-gpt3-prompt}
Prompt to generate non-figurative versions of the figurative contexts from GPT-3 and ChatGPT.}
\end{figure}

\section{Appendix B: Prompts for GPT-3 and ChatGPT for QA}
\label{sec:appendixB}
For GPT-3, we use the da-vinci-003 model with temperature set to 0 and max-length set to 1. For ChatGPT, we use gpt-3.5-turbo. In each case, we use a prompt with 2 examples, as shown in~\autoref{fig-gpt3-prompt-qa}.

\begin{figure}[htbp!]
\begin{tabular}{p{0.45\textwidth}}
\hline
Answer the following question with a yes or no based on the passage. \\ \\
Passage: The chocolate cake was sinfully decadent. \\
Question: Did the cake taste good? \\
Answer: Yes \\ \\
Passage: The camera in the phone freezes every few minutes \\
Question: Does the camera work well? \\
Answer: No \\
\hline
\end{tabular}
\caption{\label{fig-gpt3-prompt-qa}
Prompt to get yes/no answers from GPT-3 and ChatGPT.}
\end{figure}

\section{Appendix C: Chain of Thought Prompting ChatGPT for QA}
\label{sec:appendixC}
We use the gpt-3.5-turbo model. We used a prompt with 2 examples, as shown in~\autoref{fig-gpt3-prompt-cot}.

\begin{figure}[htbp!]
\begin{tabular}{p{0.45\textwidth}}
\hline
Generate a simplified version of the passage and then answer the following question with a yes or no based on the meaning of the passage. \\ \\
Passage: The chocolate cake was sinfully decadent. \\
Question: Did the cake taste good? \\
Simplified Passage: The chocolate cake was rich and delicious. \\
Answer: Yes \\ \\
Passage: The camera in the phone freezes every few minutes. \\
Question: Does the camera work well? \\
Simplified Passage: The camera stopped working every few minutes. \\
Answer: No \\
\hline
\end{tabular}
\caption{\label{fig-gpt3-prompt-cot}
Chain-of-thought prompting with ChatGPT}
\end{figure}

\section{Appendix D: Prompting ChatGPT to generate Synthetic Question Answer pairs from figurative and non-figurative contexts}
\label{sec:appendixD}
We use the gpt-3.5-turbo model. We used a prompt with 4 examples, as shown in~\autoref{fig-gpt3-syntheticQA}.

\begin{figure}[htbp!]
\begin{tabular}{p{0.45\textwidth}}
\hline
From the following text, generate a yes/no question that requires understanding the literal meaning of the text, and an answer. Refer to the examples provided. \\ \\
Text: She was a peacock in everything but looks. \\
Question: Was she pretty? \\
Answer: No \\ \\
Text: They seemed to have spared no chilli peppers in the sauce. \\
Question: Was the sauce hot? \\
Answer: Yes \\ \\
Text: The chicken was well-cooked and flavorful. \\
Question: Did the chicken taste good? \\
Answer: Yes \\ \\
Text: The pearls in the studs sparkled like the moon. \\
Question: Were the earrings dull? the? \\
Answer: No \\ \hline
\end{tabular}
\caption{\label{fig-gpt3-syntheticQA}
Prompt to generate question answer pairs from ChatGPT}
\end{figure}

\section{Appendix E: Data Statistics for Synthetic Training Data}

\autoref{tab-stats-synthetic-data} shows the distribution of synthetic training data generated from ChatGPT for the task of question answering from figurative and non-figurative contexts. 

\begin{table}[htbp!]
\centering
\begin{tabular}{c|c|c|c}
\hline
Domain & \textbf{Yes} & \textbf{No} & \textbf{Total}\\ \hline
\textbf{Yelp} & 1270 & 484 & 1754 \\ \hline
\textbf{Amazon}  & 3320 & 2102 & 5422 \\ \hline
\end{tabular}
\caption{\label{tab-stats-synthetic-data}
Distribution of yes/no questions generated by prompting ChatGPT}
\end{table}

\end{document}